\documentclass[letterpaper]{article} 
\usepackage{aaai25}  
\usepackage{times}  
\usepackage{helvet}  
\usepackage{courier}  
\usepackage[hyphens]{url}  
\usepackage{graphicx} 
\usepackage{natbib}  
\usepackage{caption} 
\usepackage{algorithm}
\usepackage{algorithmic}
\usepackage{amsmath,amssymb}
\usepackage{mathrsfs}
\usepackage{multirow}
\usepackage{subfig}
\usepackage{tabularx}
\usepackage{newfloat}
\usepackage{listings}
\newtheorem{theorem}{Theorem}

\newtheorem{prop}{Proposition}

\DeclareCaptionStyle{ruled}{labelfont=normalfont,labelsep=colon,strut=off} 
\floatstyle{ruled}
\newfloat{listing}{tb}{lst}{}
\floatname{listing}{Listing}
\title{Exploring Gradient Subspaces: Addressing and Overcoming LoRA’s Limitations in Federated Fine-Tuning of Large Language Models}
\author {
    Navyansh Mahla\textsuperscript{\rm 1},
    Kshitij Sharad Jadhav\textsuperscript{\rm 1},
    Ganesh Ramakrishnan\textsuperscript{\rm 1}
}
\affiliations {
    \textsuperscript{\rm 1}Indian Institute of Technology Bombay\\
    210040106@iitb.ac.in, kshitij.jadhav@iitb.ac.in, ganesh@cse.iitb.ac.in
}

\lstdefinelanguage{Python}{
    morekeywords={project, update, project_back},
    keywordstyle=\bfseries,
    commentstyle=\itshape,
    morecomment=[l]{\#},
    showstringspaces=false,
}

\usepackage{bibentry}

\begin{document}

\maketitle

\begin{abstract}
Large Language Models (LLMs) have demonstrated remarkable capabilities across various domains, particularly in task generalization for both text and vision data. While fine-tuning these models can significantly enhance their performance on specific downstream tasks, it often requires high-quality data that cannot be shared due to privacy concerns. Federated Learning (FL) offers a promising solution for collaborative training without direct data sharing. However, many parameter-efficient fine-tuning strategies for LLMs in FL, particularly those based on Low-Rank Adaptation (LoRA), face limitations. In this paper, we critically analyze the convergence and performance guarantees of popular FL frameworks utilizing LoRA, highlighting its suboptimal nature due to constrained subspace learning of low-rank matrices. This limitation hinders effective fine-tuning of LLMs in federated settings. Through rigorous analytical and empirical evaluations, we demonstrate that direct weight averaging outperforms LoRA-based strategies, leading to superior performance for fine-tuned models. Our comprehensive comparison unmasks inefficiencies in LoRA approaches and underscores the advantages of direct weight aggregation. We extend our analysis to low-rank gradient-based optimizers, such as GaLore, used during local training steps. Our findings show that GaLore along with direct-weight aggregation is a more effective approach, outperforming federated LoRA methods like FlexLoRA and FFA-LoRA across both text and image modalities. While privacy remains paramount in FL discourse, our focus is on assessing performance outcomes of federated fine-tuned models and evaluating various FL frameworks from both theoretical and empirical perspectives. Our findings advocate reassessing the reliance on LoRA within FL contexts, paving the way for more efficient training methodologies.
\end{abstract}

%

\section{Introduction}
The past few years have witnessed unprecedented advancements in Large Language Models (LLMs) \cite{2005.14165, openai2023gpt, 10.1145/3597503.3639219, touvron2023llama, zeng2022glm, zhang2022opt}. These language models (LMs) are powered by Transformer \cite{1706.03762} neural network architecture. Since the transformer models have extensive pre-trained context, they exhibit enhanced generalization capabilities, providing them effective few-shot learning capabilities \cite{2005.14165}. The introduction of Vision Transformers (ViTs) \cite{dosovitskiy2021an} has extended the capabilities of transformers to process image modalities. Like their counterparts in language processing, ViTs are pre-trained on vast image datasets, enabling them to achieve a robust contextual understanding of visual content. 
Models such as CLIP (Contrastive Learning Image Pre-training) \cite{pmlr-v139-radford21a} have facilitated the integration of text and image modalities by aligning their representations into a shared subspace. This has led to the development of Vision-Language Models (VLMs) and Large Multimodal Model (LMM) \cite{liu2023visual} architectures, enabling transformer networks to interpret and reason over text and image prompts simultaneously, thereby enhancing their capabilities in visual question answering. These pre-trained language models can be further fine-tuned to improve performance on specific downstream tasks \cite{pmlr-v139-radford21a}. However, the good-quality datasets required for fine-tuning can be distributed and may not be shared directly due to privacy concerns.
Researchers have turned to Federated Learning (FL) \cite{pmlr-v54-mcmahan17a} as a means to fine-tune LLMs without compromising the data privacy \cite{qin2024federatedfullparametertuningbillionsized, 2305.05644, bai2024federatedfinetuninglargelanguage, babakniya2023slora}. In these settings, parameter-efficient fine-tuning methods \cite{Ding2023} like LoRA \cite{2106.09685} are utilized to minimize computational overhead.
We analyze the most recent SOTA LoRA approaches, which include FlexLoRA \cite{bai2024federatedfinetuninglargelanguage} and FFA-LoRA \cite{sun2024improving}, and identify potential bottlenecks resulting in their performance degradation.
\newline
We demonstrate that the direct weight aggregation strategy effectively addresses the limitations of LoRA in federated learning contexts to some extent. Building on this analysis, we theoretically establish that gradient low-rank optimizers, such as GaLore \cite{zhao2024galore} coupled with direct weight averaging, represent a more effective fine-tuning approach for both large language models (LLMs) and vision transformers (ViTs) in federated settings. Our theoretical insights are validated through experiments that reveal the suboptimal performance of LoRA-based methods in FL environments. Notably, combining direct weight aggregation with GaLore as an optimizer for local training steps significantly outperforms leading state-of-the-art LoRA methods like FlexLoRA and FFA-LoRA. We list the contributions of our paper below:

\begin{itemize}
\item We highlight the sub-optimal nature of the most recent LoRA-based SOTA methods like FlexLoRA \cite{bai2024federatedfinetuninglargelanguage} and FFA-LoRA \cite{sun2024improving} in  FL.
\item We provide analytical evidence that LLMs and ViTs fine-tuned with the GaLore optimizer during local training, when directly aggregated with FedAvg, outperform methods like LoRA. Our study demonstrates this through a straightforward experimental setup using the GaLore optimizer for parameter updates and direct weight aggregation using FedAvg.
\item We present results on text and image modalities using vision and language transformers across multiple clients and configurations, surpassing current SOTA LoRA methods and validating our theoretical analysis of LoRA's sub-optimality.
\end{itemize}


\section{Related Works}
\subsection{Parameter Efficient Fine-Tuning of LLMs}
Pre-trained transformers, trained on extensive text or image datasets, gain broad contextual understanding. Fine-tuning these models improves their performance on targeted downstream tasks \cite{radford2021learning}. However, fine-tuning on consumer-grade GPUs is challenging due to their large parameter count, which demands significant GPU memory. To address this computational complexity, researchers have proposed Parameter Efficient Fine-Tuning (PEFT) \cite{Ding2023} techniques like prompt tuning \cite{2104.08691}, adapter tuning \cite{, 2106.03164, 1902.00751}, and Low-Rank Adaptation (LoRA) \cite{2106.09685}. Adapter learning techniques add trainable parameters to the model sequentially while keeping other components frozen. This reduces the overall number of trainable parameters, enabling fine-tuning on smaller GPUs. In LoRA, adapters are applied in parallel and decomposed into lower-rank matrices, further optimizing parameter efficiency. Similarly, prefix and prompt tuning involve applying modules in parallel to attention heads \cite{2101.00190} or embeddings \cite{2104.08691} to achieve efficient fine-tuning. The model's fine-tuning performance greatly depends on the dataset's size and quality \cite{Sun2024DialinsightFL}. Good-quality datasets are often distributed across multiple parties, as a single source may be insufficient for effective fine-tuning. Privacy concerns and regulations like the General Data Protection Regulation (GDPR) further prevent centralizing data for collaborative efforts, especially under the new EU Act \cite{woisetschläger2024federatedlearningprioritieseuropean}.
\subsection{PEFT in Federated Learning}
Federated Learning (FL) \cite{pmlr-v54-mcmahan17a} addresses this issue by enabling collaborative neural network training without data centralization. It has been successfully applied to various architectures, including transformers \cite{bai2024federatedfinetuninglargelanguage, qin2024federatedfullparametertuningbillionsized, 2305.05644, babakniya2023slora, kuang2023federatedscopellmcomprehensivepackagefinetuning, cho2023heterogeneous, 2110.07560, 2212.01548, niu2022federated, 2010.01264}. However, the massive size of large language models demands substantial resources for inter-client communication and local training, highlighting the need for more compute and communication-efficient paradigms in FL fine-tuning.
Due to their efficiency in fine-tuning, PEFT approaches are well-suited for FL schemes to solve such problems. LoRA has been prominently utilized in fine-tuning language models in FL settings as discussed in several studies \cite{bai2024federatedfinetuninglargelanguage, qin2024federatedfullparametertuningbillionsized, 2305.05644, cho2023heterogeneous, kuang2023federatedscopellmcomprehensivepackagefinetuning, babakniya2023slora}. Many of these studies adopt FedAvg \cite{pmlr-v54-mcmahan17a} as the aggregation algorithm to aggregate client-side parameters onto a server model. For instance, FedIT \cite{2305.05644} fine-tunes models in decentralized settings by sharing and aggregating LoRA matrices (let $\boldsymbol{B}$ be the zero-initialized LoRA matrix and $\boldsymbol{A}$ be the Gaussian initialized LoRA matrix) separately. SLoRA \cite{babakniya2023slora} introduces a two-stage sparse fine-tuning approach with improved LoRA matrix initialization for federated learning. Following the improved initialization, a mechanism similar to that of FedIT is adopted for distributed training. FlexLoRA \cite{bai2024federatedfinetuninglargelanguage} enhances previous methods by allowing diverse LoRA weight mixtures across clients, claiming superior performance compared to SLoRA in homogeneous settings and HETLORA \cite{cho2023heterogeneous} in heterogeneous settings. Instead of aggregating LoRA matrices separately, FlexLoRA multiplies matrices $\boldsymbol{B}$ and $\boldsymbol{A}$ before aggregation, then decomposes the resulting matrix into low-rank components via truncated SVD, with the decomposed matrices copied to the LoRA matrices $\boldsymbol{B}$ and $\boldsymbol{A}$ of each client. Other strategies, such as FFA-LoRA \cite{sun2024improving}, focus on fine-tuning only matrix $\boldsymbol{B}$ while keeping all other parameters frozen across varying tasks, hyperparameters, and privacy protection levels. In contrast, our approach FedFTG only fine-tunes the MLP layers without any adapters with GaLore \cite{zhao2024galore} optimization to reduce memory usage for optimization states.
\subsection{GaLore in LLMs and ViTs}
GaLore is a subspace gradient learning approach designed for memory-efficient training of LLMs. This method reduces the memory footprint of optimizer states by projecting gradient matrices into a low-rank subspace before applying optimization techniques such as AdamW \cite{1711.05101} or SGD \cite{ruder2017overviewgradientdescentoptimization}. Recent studies, such as MedSAGa \cite{mahla2024medsagafewshotmemoryefficient}, have empirically validated GaLore's effectiveness for tasks like image segmentation. Authors fine-tuned ViT-based Segment Anything Model (SAM) \cite{2304.02643} using GaLore, showcasing its applicability beyond traditional language modeling tasks.

\section{Methodology}
\subsection{Problem Setting}
We frame our problem within a multi-silo environment where each silo (client) hosts the same Large Language Model (LLM) $M$ to fine-tune along with a dataset $\mathcal{D}_i=\left\{ (x_j^i,y_j^i)_{j=1}^n \right\}$ (assuming all the $N$ clients have the same sample size $n$) that is non-independent and identically distributed (non-IID) compared to datasets of other clients. $\mathcal{D}_i^T$ and $\mathcal{D}_i^E$ respectively represent the train and eval split of the dataset $\mathcal{D}_i$. A single server facilitates the global aggregation of client parameters of $N$ clients. Due to privacy concerns, these clients cannot share the data with the server for centralized training.  $\theta_i \space \forall t \in \left \{1,2,...,N \right\}$ are the model parameters at client $i$ and $\theta_g$ are the model parameters at the server. We define the distributed learning objective as:
\begin{equation}
    \underset{\theta_g}{\text{min}} \frac{1}{N}\sum_{i=1}^{N} \left( \mathcal{L}_i(\theta_g) := \underset{(x,y)\in\mathcal{D}_i^E}{\mathbb{E}} l(M(x;\theta_g),y) \right)
\end{equation}
Where $l$ is the convex loss function. Each client performs its own local training step:
\begin{equation}
    \mathcal{L}_i(\theta_i) =\frac{1}{\left| \mathcal{D}_i^T \right|}\sum_{(x,y)\in\mathcal{D}_i^T}^{}l(M(x;\theta_i),y)
\end{equation}
After some local iterations $T_{agg}$, the parameters are sent to the server for aggregation using FedAvg:
\begin{equation*}
    \theta_g := \frac{1}{\left| \mathcal{D} \right|}\sum_{i=1}^{N}\left| \mathcal{D}_i^T \right|\theta_i
\end{equation*}
Here, $\left| \mathcal{D}\right|$ is the sum of the size of the train split of the datasets of all the clients. These aggregated parameters are subsequently copied back to each client model to resume training. This allows local iterations to proceed with the model parameters $\theta_g$, which were aggregated during the latest global aggregation step. 

\subsection{Why Not LoRA?}
In this section, we analyze the use of LoRA in FL from the perspective of two of the most recent SOTA LoRA FL schemes: FlexLoRA \cite{bai2024federatedfinetuninglargelanguage} and FFA-LoRA \cite{sun2024improving}. Both of these methods have presented their own analysis of the vanilla LoRA FL scheme FedIT \cite{2305.05644}. The authors of FlexLoRA question LoRA's efficiency in highly heterogeneous tasks across different clients. They discuss the infeasibility of existing LoRA solutions like FedIT in FL settings due to the \textit{bucket effect} caused by different intrinsic ranks at each client because of heterogeneous datasets. On the other hand, authors of FFA-LoRA perform a rigorous analysis of vanilla LoRA method like FedIT in the "client-drift" \cite{pmlr-v119-karimireddy20a} scenario and the amplification of noise in FL settings with DP-SGD \cite{Abadi_2016} due to semi-quadratic structure of LoRA. With thorough analysis, both of these methods were able to outperform vanilla LoRA FL method. Here, we present our analysis of why the SOTA LoRA frameworks like FlexLoRA and FFA-LoRA are sub-optimal under multi-client FL settings.

The core idea of LoRA is to constrain the weight update on the model by a low-rank decomposition:
\begin{equation}
    \boldsymbol{W_0} + \boldsymbol{\Delta W}=\boldsymbol{W_0}+\boldsymbol{B}\boldsymbol{A}
\end{equation}
Here $\boldsymbol{W_0} \in \mathbb{R}^{d \times k}$ is the pre-trained weight matrix which is frozen during the training process. Updates are performed on $\boldsymbol{A} \in \mathbb{R}^{r \times k} $ and $\boldsymbol{B} \in \mathbb{R}^{d \times r}$. $\boldsymbol{B}$ is initialized as zero while $\boldsymbol{A}$ uses random Gaussian initialization. Here, $r \ll \text{min}(d,k)$. This reduces the number of training parameters by a factor of $\frac{d \times k}{r \times (k+d)}$ compared to full parameter fine-tuning.


\begin{prop} In FL scenarios like FlexLoRA, where parameter change matrices $\boldsymbol{\Delta W}_i$ from $N$ clients are aggregated with each client $i$ having an intrinsic rank $r_i$, the globally aggregated parameter matrix exhibits rank inflation following each global aggregation step. Specifically, in a scenario where all clients have identical ranks $r_i=r$ for simplicity, the rate of rank inflation, denoted by $\eta$ satisfies $1 \le \eta \le N$ per global aggregation step.  \end{prop}

The equality on the left holds true when the ranks of all matrices share the same subspace, a condition that becomes unattainable when data samples are non-IID across clients. Further details, including the proof of this proposition and a more comprehensive discussion, are available in the supplementary material.\newline
In FlexLoRA, the LoRA matrices $\boldsymbol{B}$ and $\boldsymbol{A}$ are multiplied and then aggregated across clients. The product of these matrices is then subjected to Singular Value Decomposition (SVD) to decompose it back to the original LoRA rank.However, this approach faces a significant challenge: after aggregation, the resulting rank of the matrix is inflated (as shown in Proposition 1), capturing more comprehensive information from all the non-IID datasets. When this aggregated matrix is reduced to a smaller dimension via SVD, it consistently maps the weights to a subspace of fixed rank—the same as the LoRA rank, which remains constant throughout training. This dimensionality reduction can create a bottleneck, as it restricts the model's ability to efficiently capture and leverage the learned local semantics from non-IID datasets. By decomposing the weights into ranks smaller than the actual rank of the aggregated matrix, valuable information may be lost, thereby limiting the effectiveness of FlexLoRA in federated learning scenarios with diverse data distributions.
\newline
For fine-tuning LLMs in FL settings using FFA-LoRA:
\begin{equation}
    \boldsymbol{\Delta W}_{agg}=\frac{1}{N}\sum_{i=1}^{N}\boldsymbol{\Delta W}_i = \frac{1}{N}\sum_{i=1}^{N}\boldsymbol{\Delta B}_i\boldsymbol{A_0}
\end{equation}
The Gaussian initialized LoRA matrix $\boldsymbol{A}$ is frozen for all the clients. Essentially, aggregation happens for the zero initialized LoRA matrices $\boldsymbol{B}_i \forall i \in \left\{1,2,...,N  \right\}$. 
\begin{theorem}
    For a convex loss $\mathcal{L}$, let $\boldsymbol{\Delta W^*} \in \mathbb{R}^{d \times k}$ be the optimal LoRA parameter matrix, $\alpha$ be the learning rate and $\boldsymbol{A_0} \in \mathbb{R}^{r \times k}$ be a Gaussian initialized random matrix, where $r \ll \text{min}(d,k)$ and the L2 norm of the gradient to be bounded (i.e. $\left\| \boldsymbol{\nabla_W \mathcal{L}^{(i)}(\Delta W)} \right\|_2 \le D$). The excess risk ($\left| \mathcal{L}(\boldsymbol{\Delta W_{agg}}) - \mathcal{L}(\boldsymbol{\Delta W^*}) \right|$) bounds for the FFA-LoRA framework, involving $N$ clients and $S$ global aggregation steps having occured every $t_{agg}$ local training iterations, can be expressed as follows: 
    \begin{equation}
    \begin{split}
     \le DNSt_{agg}\left( DNSt_{agg}c+\frac{\alpha}{N}\left\| \boldsymbol{\Delta W^*} \right\|_2 \right) \\ = \mathcal{O}(N^2S^2t_{agg}^2)
     \end{split}
    \end{equation}

    Here, $\boldsymbol{\Delta W^*}$ is hypothetically the most optimal LoRA adapter matrix and $c$ is a constant scalar.
\end{theorem}

Theorem 1 is crucial to our analysis as it demonstrates that the excess risk bounds for FFA-LoRA are upper-bounded by an expression that increases with each FedAvg global aggregation step 
$S$. This indicates that models trained using the FFA-LoRA framework progressively deviate from the optimal hypothesis as the number of FedAvg steps increases, leading to instability. Consequently, due to this divergence, the model struggles to capture the overall data distribution across all clients, resulting in poor generalization over unseen data samples. We refer to the supplementary material for the proofs and detailed discussions on the theorem.


\begin{figure}[t]
\centering
\includegraphics[width=0.9\columnwidth]{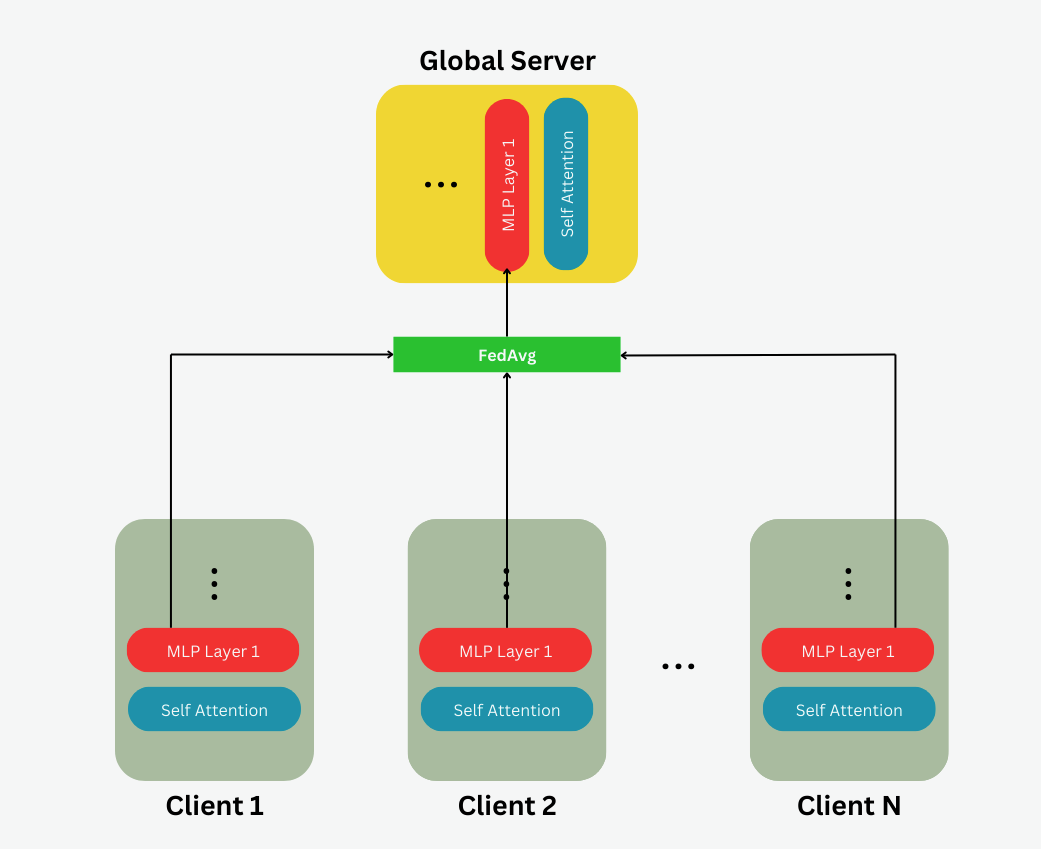} 
\caption{FedFTG exclusively fine-tunes the lower MLP layers of the transformer network while keeping all other components frozen. GaLore is used as an optimizer, and similar to standard federated learning setups, globally aggregated parameters are copied back to each client after each global aggregation round. }
\label{fig-fedftg-final}
\end{figure}

\subsection{How can we improve?}
The previous section offers valuable insights into the suboptimal behavior of LoRA in federated settings. In particular, the use of low-rank adapters hinders efficient learning, as the low-rank subspace expands with each global FedAvg aggregation step. This is particularly suboptimal for FlexLoRA, while for FFA-LoRA, the bounds on excess risk are quadratic. Therefore, we recommend against using low-rank adapters. Instead, we advocate for the direct averaging of parameters in federated settings. Direct weight aggregation in itself poses problems such as computational inefficiency. LoRA is a parameter efficient approach which results saving a lot of compute. Direct weight averaging won't be parameter efficient. To optimize memory usage and enable all clients to fine-tune LLMs locally, we recommend using GaLore as the optimizer during local training iterations. GaLore is an optimization method that improves memory efficiency while training of transformer based models \cite{zhao2024galore}. GaLore reduces optimizer state memory usage by projecting weight gradients onto a lower-dimensional subspace, where the optimization process is then carried out. Interestingly, this approach not only saves memory but also leads to linear excess risk bounds, as demonstrated in the following theorem:

\begin{theorem}
    For a convex loss function \(\mathcal{L}\), let \(\boldsymbol{\Delta W^*}\) denote the optimal weight matrix and \(\alpha\) represent the learning rate. Assuming that the L2 norm of the gradient is bounded, specifically \(\left\| \boldsymbol{\nabla_W \mathcal{L}^{(i)}(\Delta W)} \right\|_2 \le D\), the excess risk, defined as \(\left| \mathcal{L}(\boldsymbol{\Delta W_{agg}}) - \mathcal{L}(\boldsymbol{\Delta W^*}) \right|\), for the aggregated weights after a total of $S$ direct FedAvg aggregations having occured every \(t_{agg}\) local training iterations with GaLore as an optimizer can be expressed as follows:

\begin{equation}
   \left| \mathcal{L}(\boldsymbol{\Delta W_{agg}}) - \mathcal{L}(\boldsymbol{\Delta W^*}) \right| \le \alpha D^2St_{agg}+c = \mathcal{O}(St_{agg})
\end{equation}

Here, $c$ is a scalar constant.
\end{theorem}

Concluding from theorem 2, the upper bounds on excess risk for direct weight averaging are independent of the number of clients and exhibit a linear relationship (contrary to quadratic in case of LoRA (theorem 1)) with the number of global FedAvg steps and local training iterations \( t_{agg} \). Adopting direct weight averaging with GaLore as an optimizer for fine-tuning in federated settings ensures that excess risk remains unaffected by client count. This trend is further supported by the experimental results presented in later sections. We will use FedAvg as the global aggregation algorithm, as it is widely used in federated fine-tuning for large language models. Combining FedAvg with direct weight averaging allows for a fair comparison and shows that even a simple algorithm like FedAvg can yield significantly better results when paired with a more efficient framework. 
In the context of transformer neural networks, the GaLore paper demonstrates that the low-rank projection of gradient matrices can be effectively applied to its lower MLP layers like \textit{project-up} (see Lemma B.6 from GaLore paper) whose gradients become low-rank during training. Consequently, unless stated otherwise, we conduct all our analyses and experiments on the lower MLP layers. Another advantage of using GaLore is its improved generalization error compared to LoRA-based methods, as demonstrated by the following theorem (None of the theorems assume a specific client dataset size; they account for both equal and different dataset sizes):

\begin{theorem}
\label{th: theorem-3}

    Let $N$ denote the number of clients, each possessing a dataset with $n$ samples. We consider the weights of a lower-level MLP layer represented by $\boldsymbol{W}\in \mathbb{R}^{d\times k}$. Under the assumption that the risk function of each client is $\sigma$-sub-Gaussian with respect to the data distribution of that client and the corresponding weight matrix, we derive the following generalization error bounds for weight aggregation:

    a) For federated fine-tuning using FFA-LoRA:

    \begin{equation}
       \mathcal{E}_1\le \frac{1}{N}\sum_{i=1}^N\sqrt{\frac{2\sigma^2 \ln2}{n}rq\sum_i(d)}
    \end{equation}

    b) or direct weight aggregation using FedAvg with a low-rank gradient-based optimizer, such as GaLore:
    \begin{equation}
        \mathcal{E}_2\le\frac{1}{N}\sum_{i=1}^N\sqrt{-\frac{2\sigma^2}{n}\sum_{j=1}^d\sum_{l=1}^k f(\boldsymbol{W_i^{(j)}[l]}, t)\log\left( f(\boldsymbol{W_i^{(j)}[l]}, t) \right)}
    \end{equation}
    where $\boldsymbol{W_i^{(j)}[l]}$ represents the element of $l$th index in the $j$th row of the matrix $\boldsymbol{W_i}$. Here, $f(\boldsymbol{W_i^{(j)}[l]}, t)=\frac{ \exp{(\boldsymbol{W_i^{(j)}[l]}})}{\sum_{l=1}^k \exp{(\boldsymbol{W_i^{(j)}[l]}}) }$ be the function to represent the ratio (probability) in a succinct manner. As $t \rightarrow \infty $, i.e. as the federated training continues $f(\boldsymbol{W_i^{(j)}[l]}, t) \rightarrow \frac{1}{k}$.
\end{theorem}

As one can infer from the theorem, generalization error $\mathcal{E}_2$ at the end of the training becomes upper bounded by $\frac{1}{N}\sum_{i=1}^{N}\sqrt{\frac{2\sigma^2 d}{nk}\log{k}}$ which is much smaller than that of the part (a) ($\mathcal{E}_1\le \frac{1}{N}\sum_{i=1}^{N}\sqrt{\frac{2\sigma^2 \ln 2}{n}rqd}$).
Thus, direct averaging with GaLore optimizer shows better generalization than FFA-LoRA framework. Notably, overall entropy decreases during GaLore-based direct FedAvg aggregation, indicating more structured weight matrices, which correlates with enhanced generalization performance. Integrating the GaLore optimizer with FedAvg effectively leads to rank reduction (considering fixed quantization bits $q$), confining the learning process to a compact, structured subspace with lower entropy. This localized learning is important as it achieves comparable excess risk bounds while operating in a reduced parameter space, enhancing computational efficiency. Our analysis shows salient feature learning in federated settings similar to \cite{tian2024joma} which is for centralized cases. The framework focuses on learning shared salient features across distributed datasets, creating a common subspace that facilitates effective performance across all client distributions. As training advances, the parameter matrix converges to a well-defined, approximately linear manifold due to the reduced rank. The gradient matrix exhibits similar low-rank behavior as $t \rightarrow \infty$, suggesting it also manifests on a near-linear manifold. GaLore integration in federated learning outperforms traditional LoRA-based weight averaging through better efficiency, better risk bounds, and stronger generalization guarantees. This leads us to explore optimal ways to implement GaLore for distributed fine-tuning to maximize its subspace learning benefits. To address this challenge, we introduce \textbf{FedFTG} (\textbf{Fed}erated \textbf{F}ine \textbf{T}uning using \textbf{G}aLore), an experimental framework for federated learning that is designed for federated fine-tuning scenarios while using GaLore for local training steps at each client. 


\begin{figure}[t]
\centering
\includegraphics[width=0.45\textwidth]{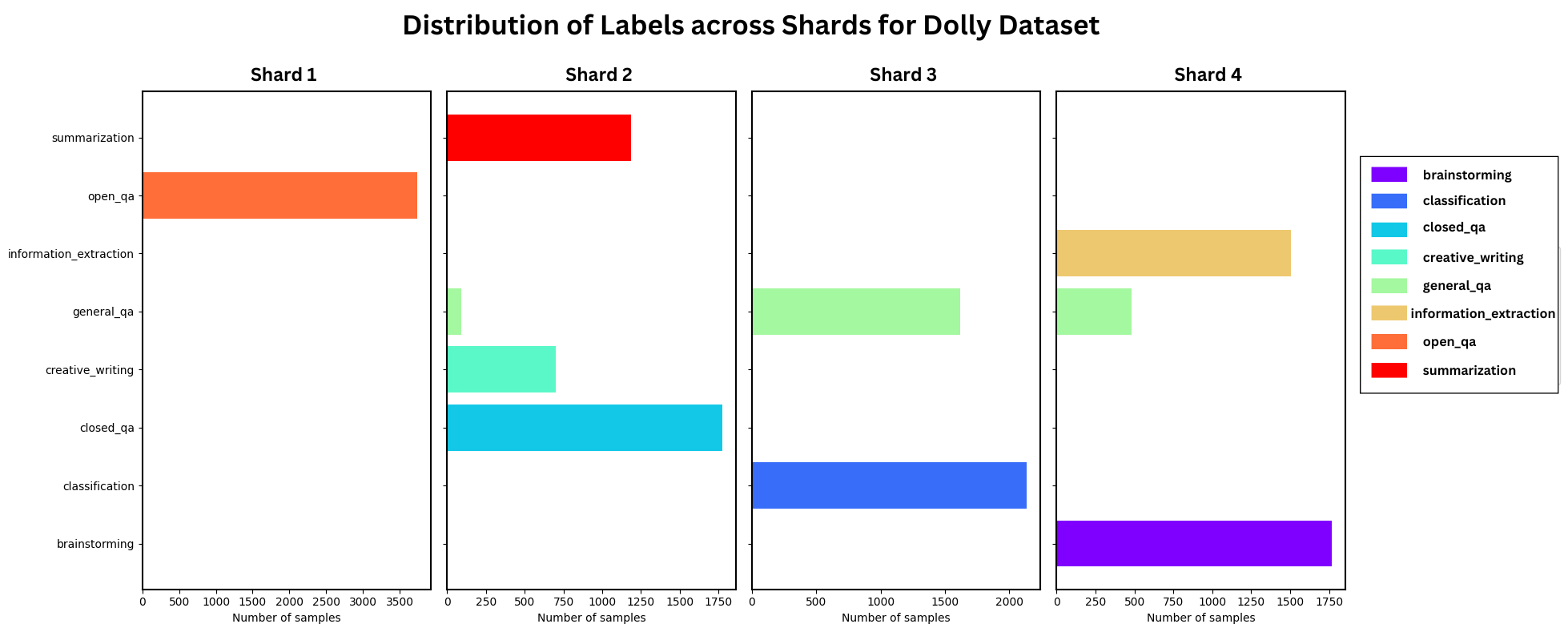}
\caption{Label distribution across shards for the Dolly dataset produced using Dirichlet Allocation with $\alpha=0.1$.}
\label{fig-sharddolly}
\vspace{0.5cm} 
\includegraphics[width=0.45\textwidth]{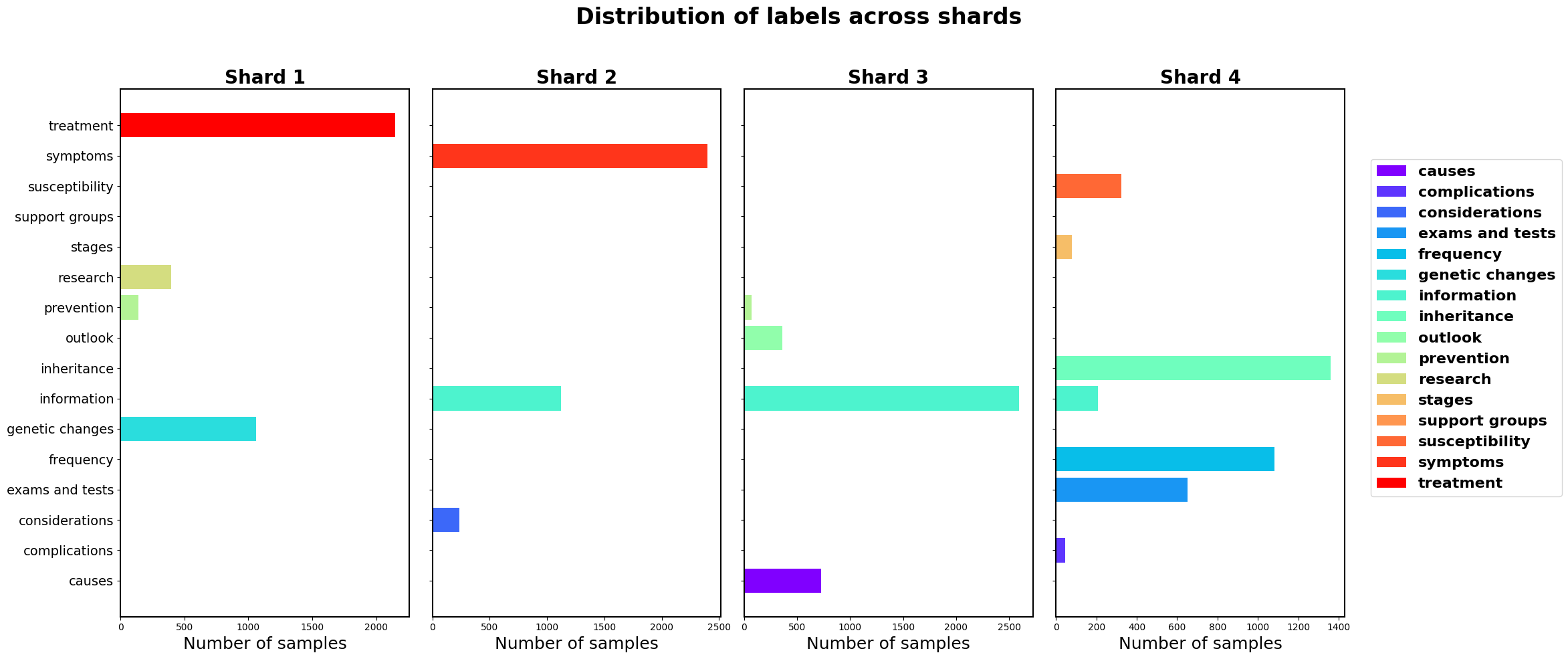}
\caption{Label distribution across shards for the MedQuAD dataset produced using Dirichlet Allocation with $\alpha=0.1$.}
\label{fig-shardmedquad}
\end{figure}

\subsection{Federated Fine-Tuning Using GaLore (FedFTG)}

In previous sections, we highlighted the limitations of low-rank adapter tuning for LLMs in federated settings. We also demonstrated that using GaLore as an optimizer for local training offers superior generalization, improved memory efficiency, and better training performance overall. We refer to the JoMA paper \cite{tian2024joma} (Theorem 1 of the paper) that states that we do not need to explicitly update the self-attention parameters since it is already implicitly incorporated in the lower layer of MLP weight. Consequently, we focus on fine-tuning the lower MLP layers (\textit{project-up}) of the transformer neural network. This is consistent with Theorem 3, which addresses the scenario where the lower MLP layers are fine-tuned. Since the self-attention information is inherently captured within these lower MLP layers, and these layers typically exhibit lower generalization error in federated settings, we recommend focusing on fine-tuning only the lower MLP layers. To update the parameters in FedFTG, we utilize GaLore, as detailed in the previous section. The entire pipeline is shown in the figure \ref{fig-fedftg-final}. The complete FedFTG algorithm pipeline is detailed in the accompanying algorithm. FedFTG is an experimental setup that validates our theory against current SoTA LoRA approaches, addressing their low-rank learning constraints and training stability issues.

\begin{algorithm}[tb]
\caption{FedFTG}
\label{alg:FedFTG}
\textbf{Input}: Model $\mathcal{M}$ from each client $i$ with non-IID datasets $\mathcal{D}_i = \left\{ (x_j^i, y_j^i) \right\}_{j=1}^n$, learning rates $\eta_i$\\
\textbf{Parameter}: Client parameters $\theta_i$, global parameters $\theta_g$, aggregation period $T_{agg}$, total mini-batches $T$\\
\textbf{Output}: Optimized global parameters $\theta_g$ after $e$ epochs
\begin{algorithmic}[1]
\STATE Initialize $\theta_g$
\FOR{epoch $= 1,\ldots,e$}
    \STATE Initialize all local $\theta_i \gets \theta_g$ in parallel
    \FOR{$t = 1,\ldots,T$}
        \FORALL{client $i \in \{1,\ldots,N\}$ in parallel}
            \STATE Sample mini-batch $\mathcal{B}_i$ from $\mathcal{D}_i$
            \STATE $G(\theta_i, \mathcal{B}_i) \gets \nabla_{\theta_i} \mathcal{L}(\mathcal{M}(\theta_i; \mathcal{B}_i), \mathcal{B}_i)$
            \STATE $\theta_i \gets \theta_i - \eta_i \text{GaLore}(G(\theta_i, \mathcal{B}_i))$
        \ENDFOR
        \IF{$t \bmod T_{agg} = 0$}
            \STATE $\theta_g \gets \frac{1}{N} \sum_{i=1}^{N} \theta_i$
            \STATE Update all local $\theta_i \gets \theta_g$ in parallel
        \ENDIF
    \ENDFOR
\ENDFOR
\RETURN $\theta_g$
\end{algorithmic}
\end{algorithm}

\section{Experiments}
In this section, we demonstrate the stability and efficiency of models trained with GaLore as the optimizer and direct weight aggregation. We evaluate its performance across both text and image modalities to showcase its effectiveness on LLMs and ViTs. For the text modality, experiments involve 3 and 4 clients, while vision experiments include 3, 4, and 5 clients, with each client hosting a non-IID dataset distinct from others. All experiments were conducted on a cluster of Nvidia A6000 GPUs, with setup details provided in the supplementary material.

\subsection{Datasets}
We conduct experiments on both text and image datasets, incorporating various downstream tasks. For text, we use the MedQuAD \cite{BenAbacha-BMC-2019} and Databricks Dolly 15k \cite{DatabricksBlog2023DollyV2} datasets. MedQuAD is a medical question-answering dataset containing 47,457 question-answer pairs sourced from 12 NIH websites, covering 39 question types related to diseases, drugs, and other medical entities. However, owing to the MedlinePlus copyright, answers from the 3 subsets were removed. Databricks Dolly-15k contains 15,000 high-quality human-generated prompt/response pairs designed for instruction tuning LLMs. It includes various categories like brainstorming, classification, summarization, and question answering. For experiments related to vision modality, we utilize the Brain Tumour classification dataset \cite{https://doi.org/10.6084/m9.figshare.1512427.v5} which comprises of 3,064 T1-weighted contrast-enhanced MRI images from 233 patients, categorized into three tumor types: meningioma (708 slices), glioma (1,426 slices), and pituitary tumor (930 slices). The task is to identify the correct tumor type by having the MRI image fed as an input to the ViT. More details on these datasets are discussed in the data appendix of the supplementary material.

\subsection{Non-IID Data Preparation}
To simulate non-IID conditions, we used Dirichlet Allocation to partition each dataset into several non-IID splits similar to \cite{2305.05644, hsu2019measuringeffectsnonidenticaldata}. For the text datasets (MedQuAD and Dolly 15k), we generated 4 splits, whereas for the Brain Tumor Classification dataset, we created 5 splits. The concentration parameter 
$\alpha$ was set to 0.1, resulting in highly skewed distributions across clients that replicate non-IID behavior in FL settings. Splits are created based on labels. For MedQuAD dataset, the label is \textit{question\_type} while for Dolly-15k it is \textit{category}. The shards prepared for a given dataset are of the same sizes. Class label distribution across each non-IID shard for all the datasets (both image and text) can be found in the supplementary material.



\subsection{Models and Hyperparameters}
For experiments on text datasets (MedQuAD and Dolly-15k), we utilize Gemma-2B \cite{gemmateam2024gemmaopenmodelsbased} and TinyLlama \cite{zhang2024tinyllamaopensourcesmalllanguage}. In FedFTG, we fine-tuned the \textit{up-proj} MLP layer, which follows the self-attention module. Conversely, FlexLoRA and FFA-LoRA focused on fine-tuning the attention modules (query, key, value) with a LoRA rank of 8 and a scaling factor of 16, in line with the experimental setups presented in their original papers. For vision modality experiments, we fine-tuned SigLIP \cite{2303.15343}. In FedFTG applied to SigLIP, we fine-tuned the MLP layer immediately after the attention module and the classifier layer (see supplementary material for more details). FlexLoRA and FFA-LoRA, on the other hand, fine-tuned the attention parameters (query, key, value) with a LoRA rank of 8 similar to text experiments and a scaling factor of 32.

\subsection{Training and Evaluation}
We conducted our experiments with different numbers of clients. For text datasets, we evaluated FedFTG with 3 and 4 clients. With 3 clients, shards 1, 2, and 3 of each dataset were used, while with 4 clients, all shards were utilized. A similar shard assignment was used for the Brain Tumour Classification dataset. Each shard was split into training and test sets, with 1\% of samples reserved for testing in the text datasets and 5\% for the vision dataset (vision dataset shards are much smaller thus larger fraction of test set is required for making a firm conclusion). The global evaluation set was created by combining the test sets from each shard, thus containing unseen samples from different non-IID shards. This helps assess the model's generalization ability in distributed non-IID settings. Text datasets were trained with a batch size of 1, while the image dataset used a batch size of 2, with all models trained for 3 epochs.

\subsection{Experiment Results}
\begin{figure}[t]  
\centering
\subfloat[3 clients, MedQuAD, TinyLlama]{
    \includegraphics[width=0.45\columnwidth]{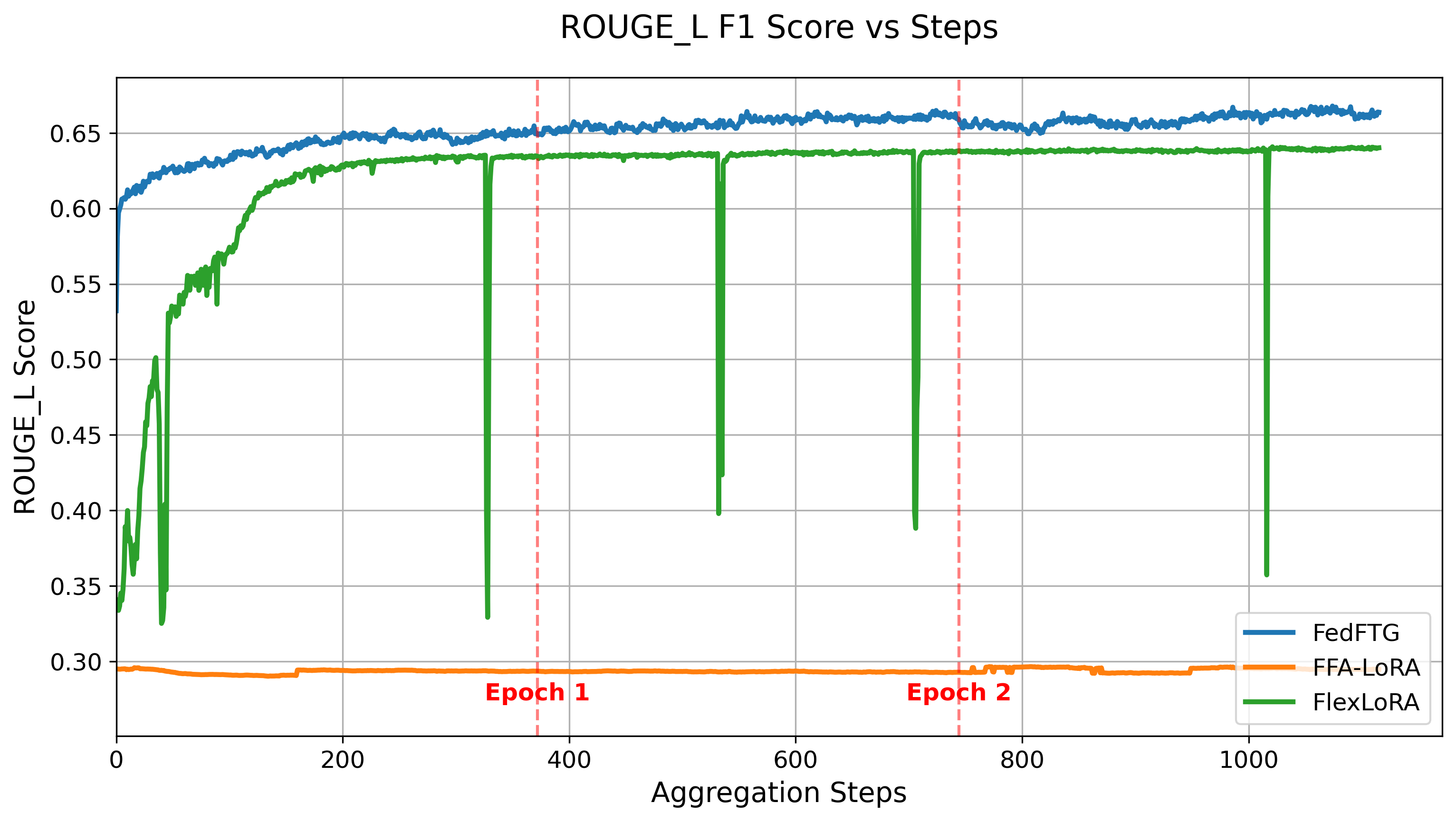}  
    \label{fig-agg3ctiny}
}
\hfill
\subfloat[4 clients, MedQuAD, TinyLlama]{
    \includegraphics[width=0.45\columnwidth]{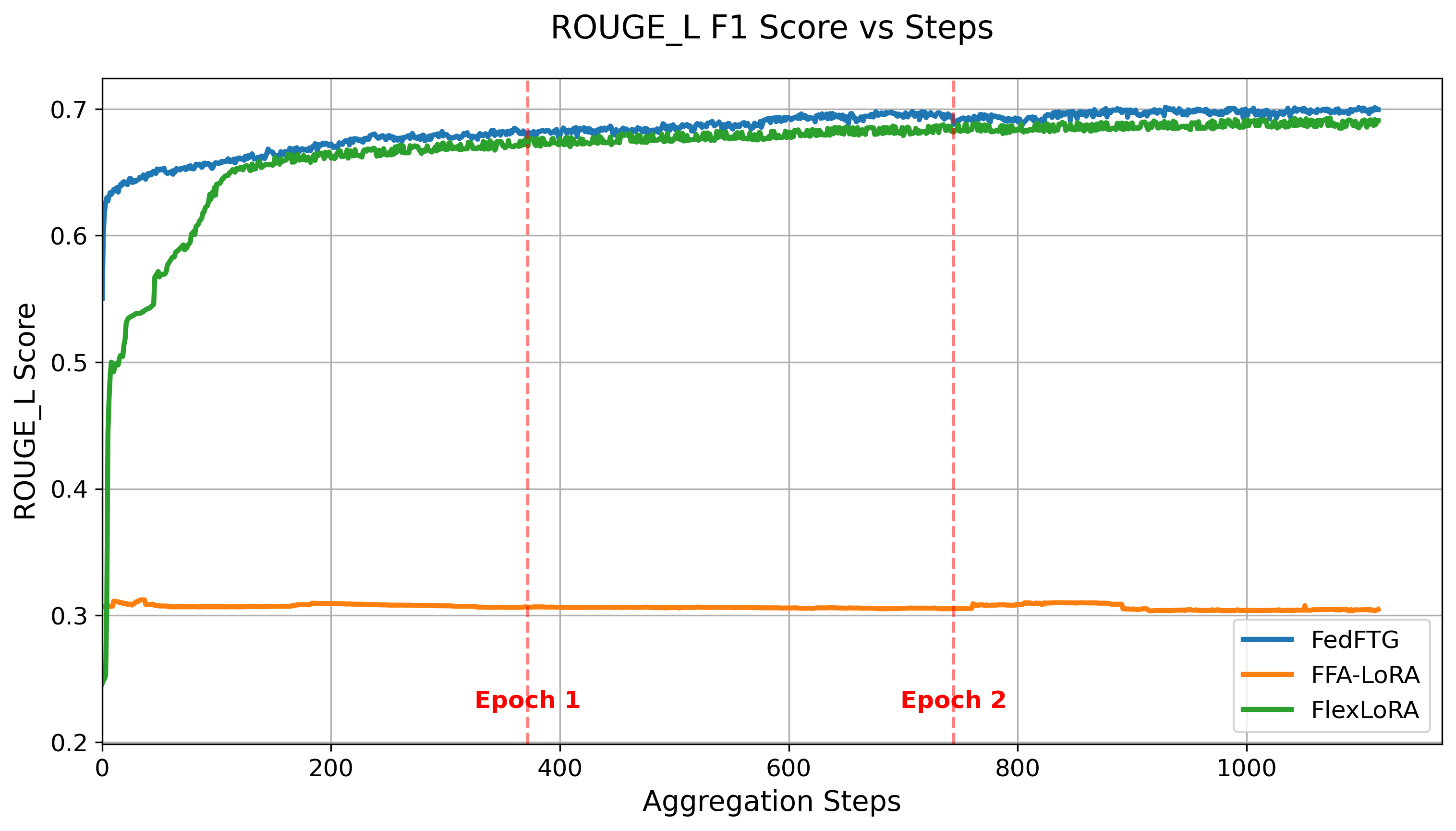}  
    \label{fig-agg4ctiny}
}
\vspace{0.05cm} 
\subfloat[3 clients, Dolly, TinyLlama]{
    \includegraphics[width=0.45\columnwidth]{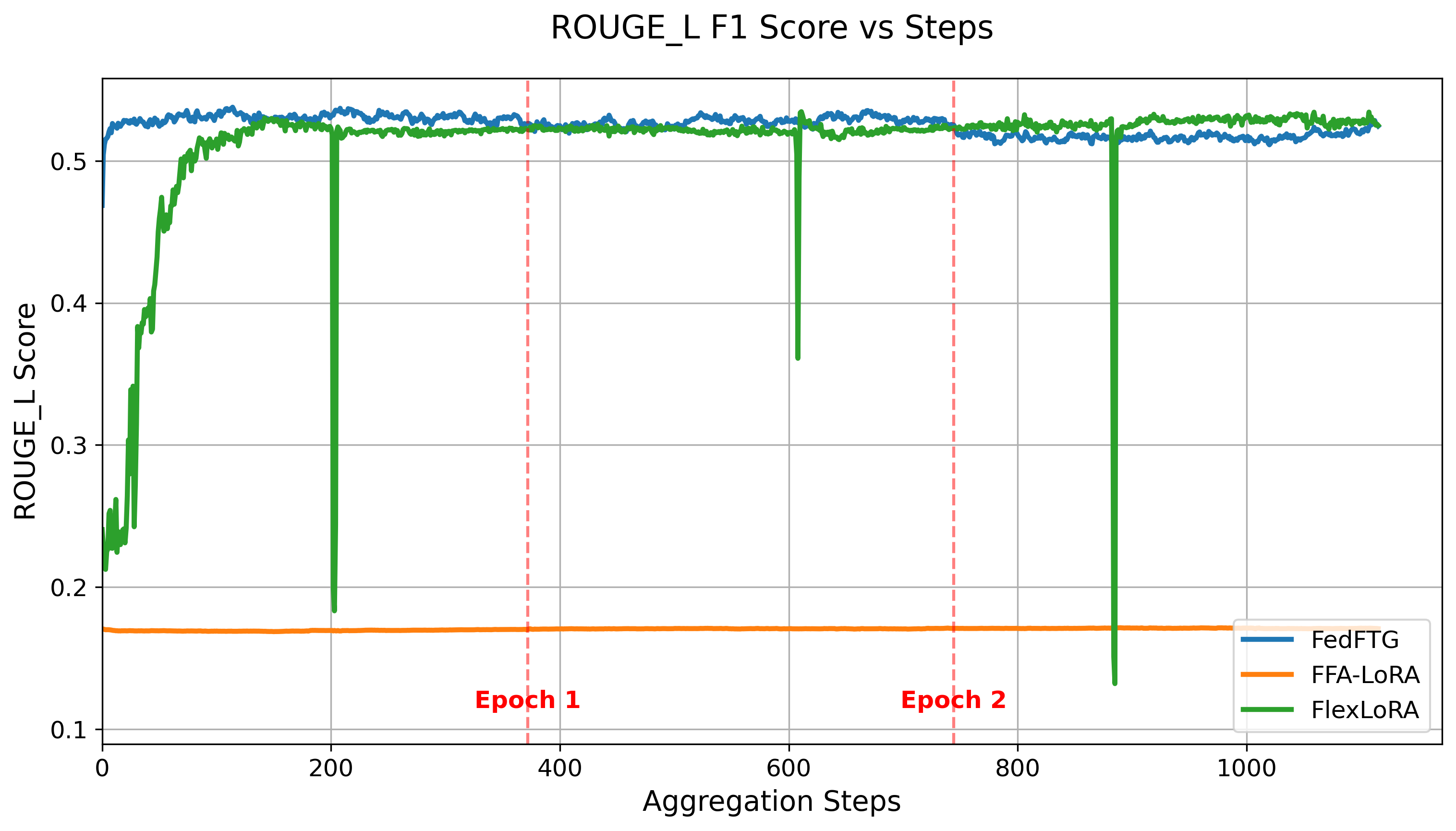}  
    \label{fig-agg3ctinydolly}
}
\hfill
\subfloat[4 clients, Dolly, TinyLlama]{
    \includegraphics[width=0.45\columnwidth]{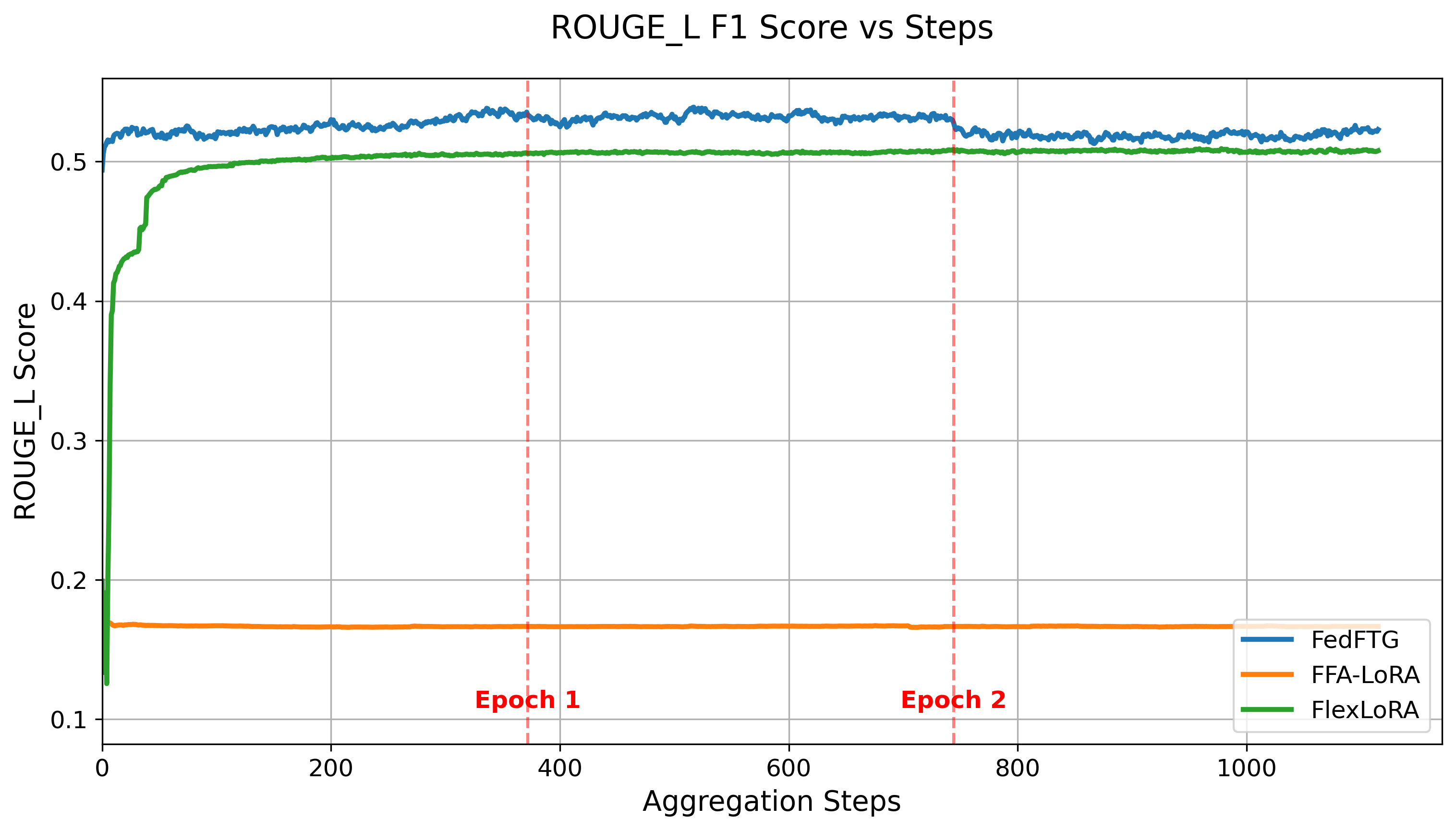}  
    \label{fig-aggadditional}
}
\caption{Variation of ROUGE\_L scores evaluated on the test set with global aggregation steps across different clients and datasets for the TinyLlama model.}
\label{fig-agg-combined}
\end{figure}

Table \ref{tab1:results-main} shows the results on text datasets. We report the ROUGE\_L F1 (Longest common subsequence ROUGE) score \cite{lin-2004-rouge} and the BLEU-4 (4-gram) \cite{papineni-etal-2002-bleu} score for text datasets. Out of 3 epochs, we report the results of the epoch which has the best result. This is done owing to the overfitting of models usually in some experiment runs. 
Table 1 shows that FedFTG consistently surpasses both FlexLoRA and FFA-LoRA across different numbers of clients and datasets.
Table \ref{tab2:results-vision} 
shows the results of fine-tuning SigLIP on the Brain Tumor Classification Dataset for tumor type classification. In this vision dataset experiment, FedFTG outperforms both LoRA-based FL methods. The notably poorer performance of FFA-LoRA across text and vision experiments aligns with earlier analyses, as its excess risk bounds increase with each FedAvg step (Theorem 1), hindering efficient learning of the dataset's semantics and leading to suboptimal results.
Figure \ref{fig-agg-combined} displays the variation in ROUGE\_L F1 scores on the global evaluation set across FedAvg steps for TinyLlama, with similar graphs for the Gemma-2B model in the supplementary material. FFA-LoRA's ineffective aggregation, shown at the bottom of the graphs, shows minimal improvements with more FedAvg steps. While FlexLoRA experiences performance dips on the MedQuAD dataset, it still achieves reasonable aggregation efficiency but falls short compared to FedFTG. In contrast, FedFTG demonstrates the most stable aggregation results, maintaining consistent training performance without abrupt drops. Overall, FedFTG outperforms both FlexLoRA and FFA-LoRA across various clients, datasets, and downstream tasks
(Table \ref{tab1:results-main} and \ref{tab2:results-vision}). \newline
The supplementary material includes more detailed explanation of the results and additional experiments with varying label distributions. Across these configurations, FedFTG consistently outperforms FlexLoRA and FFA-LoRA, with results validated over two independent runs, reinforcing our findings.

\begin{table}[]
\centering
\begin{tabular}{l|l|l|l|l|l}
\hline
$N$                 & Dataset                    & Model                      & Method   & BLEU-4          & ROUGE-L         \\ \hline
\multirow{12}{*}{3} & \multirow{6}{*}{MedQuAD}   & \multirow{3}{*}{TinyLlama} & FedFTG   & \textbf{0.4883} & \textbf{0.6637} \\
                    &                            &                            & FlexLoRA & 0.4551          & 0.6429          \\
                    &                            &                            & FFA-LoRA & 0.1004          & 0.2947          \\ \cline{3-6} 
                    &                            & \multirow{3}{*}{Gemma-2B}  & FedFTG   & \textbf{0.5493} & \textbf{0.7019} \\
                    &                            &                            & FlexLoRA & 0.4238          & 0.6401          \\
                    &                            &                            & FFA-LoRA & 0.1077          & 0.2875          \\ \cline{2-6} 
                    & \multirow{6}{*}{Dolly-15K} & \multirow{3}{*}{TinyLlama} & FedFTG   & \textbf{0.3157} & \textbf{0.5244} \\
                    &                            &                            & FlexLoRA & 0.2873          & 0.5221          \\
                    &                            &                            & FFA-LoRA & 0.0566          & 0.1708          \\ \cline{3-6} 
                    &                            & \multirow{3}{*}{Gemma-2B}  & FedFTG   & \textbf{0.3284} & \textbf{0.5413} \\
                    &                            &                            & FlexLoRA & 0.2874          & 0.5375          \\
                    &                            &                            & FFA-LoRA & 0.1077          & 0.2875          \\ \hline
\multirow{12}{*}{4} & \multirow{6}{*}{MedQuAD}   & \multirow{3}{*}{TinyLlama} & FedFTG   & \textbf{0.5433} & \textbf{0.6994} \\
                    &                            &                            & FlexLoRA & 0.501           & 0.6816          \\
                    &                            &                            & FFA-LoRA & 0.1133          & 0.3047          \\ \cline{3-6} 
                    &                            & \multirow{3}{*}{Gemma-2B}  & FedFTG   & \textbf{0.5373} & \textbf{0.7114} \\
                    &                            &                            & FlexLoRA & 0.4690          & 0.6863          \\
                    &                            &                            & FFA-LoRA & 0.1092          & 0.2863          \\ \cline{2-6} 
                    & \multirow{6}{*}{Dolly-15K} & \multirow{3}{*}{TinyLlama} & FedFTG   & \textbf{0.3428} & \textbf{0.5229} \\
                    &                            &                            & FlexLoRA & 0.2807          & 0.5081          \\
                    &                            &                            & FFA-LoRA & 0.0619          & 0.1665          \\ \cline{3-6} 
                    &                            & \multirow{3}{*}{Gemma-2B}  & FedFTG   & \textbf{0.3423} & \textbf{0.5529} \\
                    &                            &                            & FlexLoRA & 0.3061          & 0.5516          \\
                    &                            &                            & FFA-LoRA & 0.0661          & 0.1648          \\ \hline
\end{tabular}
\caption{Comparison of BLEU-4 and ROUGE L F1 scores across different methods, models, and datasets for varying client numbers with non-IID splits}
\label{tab1:results-main}
\end{table}

\begin{table}[]
\centering
\begin{tabular}{lll}
\hline
No of clients      & Method   & F1 Score        \\ \hline
\multirow{3}{*}{3} & FedFTG   & \textbf{0.6064} \\
                   & FlexLoRA & 0.5638          \\
                   & FFA-LoRA & 0.0165           \\ \hline
\multirow{3}{*}{4} & FedFTG   & \textbf{0.843}  \\
                   & FlexLoRA & 0.4868          \\
                   & FFA-LoRA & 0.3175          \\ \hline
\multirow{3}{*}{5} & FedFTG   & \textbf{0.7274} \\
                   & FlexLoRA & 0.6116          \\
                   & FFA-LoRA & 0.1104         
\end{tabular}
\caption{Comparison of F1 score on Brain Tumour Classification Dataset for fine-tuning SigLIP using different FL fine-tuning methods across non-IID splits}
\label{tab2:results-vision}
\end{table}

\section{Conclusion and Future Work}
Our analysis begins by examining LoRA's performance in federated settings, focusing on state-of-the-art frameworks FlexLoRA and FFA-LoRA, with particular attention to constraints imposed by low-rank adapter subspace learning. This investigation extends to direct weight aggregation methods and the application of GaLore as an optimization strategy, where we provide theoretical insights demonstrating its advantages over LoRA-based approaches. We introduce FedFTG, a streamlined approach for fine-tuning transformer models in federated environments that leverages GaLore's efficient subspace learning mechanism. This framework demonstrates robust performance across vision and language transformer architectures, effectively addressing LoRA's sub-optimal subspace learning bottlenecks while consistently outperforming existing approaches across diverse datasets and client configurations.
By incorporating GaLore's insights, we show how gradient subspace methods operating on parameter matrices can mitigate common federated learning bottlenecks, particularly preventing rank inflation and its associated excess risk increase. This success suggests that exploiting slowly evolving gradient subspaces could lead to more robust aggregation algorithms. Future research directions include developing more stable, parameter-efficient federated fine-tuning methods and exploring adaptive aggregation strategies for heterogeneous environments. Our work aims to guide the research community toward low-rank gradient-based optimization strategies, while the theoretical foundations established here could inform improved aggregation frameworks focusing on structured localized subspaces.
\bibliography{aaai25}

\section{Reproducibility Checklist}
All the details related to the reproducibility checklist can be found in the supplementary material.

\section{Acknowledgment}
The Authors Acknowledge the Amazon IIT-B AI ML Initiative for funding this research work.

\end{document}